# Transgender Community Sentiment Analysis from Social Media Data: A Natural Language Processing Approach

**Yuqiao Liu**
Northeast Yucai Bilingual School
Shenyang, Liaoning, China
hernandolyq@163.com

**Yudan Wang**
Northeast Yucai Foreign Language School
Shenyang, Liaoning, China
1359414853@qq.com

**Ying Zhao**
Department of Engineering Science and Applied Math
Northwestern University
Evanston, IL, U.S.A
yingzhao2018@u.northwestern.edu

**Zhixiang Li**
Department of Biomedical Engineering
Shenyang Pharmaceutical University
Shenyang, Liaoning, China
zhixiangli@gmail.com

## Abstract

Transgender community are experiencing a huge disparity in mental health condition compared with the general population. Interpreting the social medial data posted by transgender people may help us understand the sentiments of these sexual minority group better and apply early interventions. In this study, we manually categorize 300 social media comments posted by transgender people to the sentiment of negative, positive and neutral. 5 machine learning algorithms and 2 deep neural networks are adopted to build sentiment analysis classifiers based on the annotated data. Results show that our annotations are reliable with a high Cohen's Kappa score over 0.8 across all three classes. LSTM model yields an optimal performance of accuracy over 0.85 and AUC of 0.876. Our next step will focus on using advanced natural language processing algorithms on larger annotated dataset.

## 1 Introduction

Transgender is defined as a person whose gender self-identity differs from their birth sex[1]. Many transgender people experience serious gender dysphoria [2], which is a depression feeling due to the sexual mismatch between their self-identity and assigned sex. Therefore, medical treatments such as hormone replacement therapy or psychotherapy, are commonly sought by transgender people. According to research by Amnesty International [3], about 0.3% of the EU's population is transgender, accounting for 1.5 million people. In Unites States, an estimated proportion of 0.5 to 0.6% population are identified as transgender. In other countries or regions, the statistics of transgender are less reported, due to culture difference, religion reason or other reality obstacles.

According to the research of National Center for Transgender Equality (NCTE) [4], transgender community experience a huge disparity in healthcare access, employment, and criminal justice system. To be detailed, 14% of the transgender people are unemployed which is double of the rate in general population; 19% transgender people have been refused to be provided health care; approximately one-fifth of the community have reported being harassed by police. Most of these huge disparity are caused by the discrimination from people with other sexual orientations. Besides this, transgender people are also experiencing serious mental health challenges. Transgender itself is not a mental disorder. However, gender dysphoria will make people have depression and distressing feelings. Moreover, the confusion of self-identification and misunderstand from others will enhance this disorder. This mental health problem will result in unexpected health outcomes even suicide [5]. 41% of transgender individuals have attempted suicide found by the study of NCTE.

Therefore, it is essential to understand the sentiment of transgender community and give appropriate intervention or prevention to those people having gender dysphoria or other mental illness. However, transgender people, being a sexual minority, are less represented or spoken out to the public [6]. Due to the anonymous schema, people are more willing to express their own opinion to the public on the social media. To this end, social media data is a perfect data resource to the research of sentiment analysis of the transgender community. Therefore, we will conduct sentiment analysis using natural language processing approach on social media data posted by transgender community. To be detailed, our contribution is in 2-fold: 1) To give positive, negative or neural annotations to 500 contexts extracted from Reddit data posted by transgender community(sub-reddit:/r/asktransgender); 2) Implement trending natural language processing methods on the annotated data to build a automatic sentiment classifier.

Table 1: The definition, count, prevalence, example and Cohen's Kappa score of the sentiments of 300 Reddit comments posted by transgender community

| Sentiment | Defination | Count | Prevalence | Cohen's Kappa | Example |
|---|---|---|---|---|---|
| Negative | Feeling uncomfortable, depressed, unconfident of being a transgender person; Experiencing dysphoria; Being scared of receiving medical treatments and all other negative sentiments caused by transgender. | 72 | 24.00% | 0.825 | *I've really struggled trying to find a way to express how I felt about my own gender issues. I've had them for years and years now, but at the same time I've rarely showed outward evidence of femininity in my daily life ...* |
| Positive | Feeling comfortable, confident, and proud of being a transgender people; Positively receiving medical treatment; Helping other people in the community and all other positive feelings associated with transgender. | 85 | 28.33% | 0.877 | *Reading a lot about gender dysphoria and online tests re: transgender identity. I realize that while I (AMAB) have recently begun intermittently fantasizing about being a woman, I don't have any negative feelings about my life prior to these...* |
| Neutral | The posts can't be categorized to negative or positive sentiment ; Or can be categorized to both sentiments. | 143 | 47.67% | 0.902 | *So, I've got an appointment coming up with my primary care physician, basically to attempt to suss out whether I could go thru the process of getting hormones thru them. and I think I've got a basic idea of what I should be asking, but I'm not sure what to really ask or to expect.* |

## 2 Related Work

With the fast development of artificial intelligence (AI) in the last decade, these techniques has been widely applied to the research different fields. Particularly in healthcare, AI is used to improve the predictive accuracy of diseases[7, 8], to interpret medical images automatically [9, 10], to explore the RNA sequence pattern between co-morbidity [11] and so on. In terms of sentiment analysis of social media data, natural language processing is heavily used. Kanakaraj et al. [12] apply ensemble classifiers from semantic features to improve the performance of classification tasks compared with traditional bag-of-word models. Yadav et al [13] build a convolutional neural networks (CNN) -based model to implement a patient assisted system from patient.info data. Broek-Altenburg et al. ][14] analysis consumers' sentiments of purchasing health insurance during enrollment season. These techniques are also applied to the reseach of sexual minority groups' sentiment analysis using social media data. Fitri et al. [15] conduct sentiment analysis of twitter data using machine learning algorithms with focus on an anti-LGBT campaign in Indonesia. Khatua et al. [16] explores the tweeting attitude in support of LGBT community in India with a deep learning approach. Unlike previous research, we find transgender group is less discussed and may have more serious mental health concerns. To these limitations, we will apply natural language processing techniques to analyze transgender peoples' sentiments on social media.

## 3 Methods

### 3.1 Dataset

We use the dataset pulished at https://github.com/mjtat/Trans-NLP-Project. In this dataset, more than 20 thousand comments are crawled from the sub-reddit /r/asktransgender. We can not annotate the sentiments of each comment as it requires tremendous labor. Thus, we manually annotated 300 of all extracted comments and apply natural language processing to automatic annotate the other comments. Each comment will be categorized as one of the three sentiments: negative, positive and neutral. Two annotators are involved in our annotation process. They first discuss the definition of each label and then reach a consensus. The Cohen's kappa [17] score of each label is calculated to evaluate the inter-rater reliability. For those comments that both annotators can't make a consensus, annotator3 is involved to give a find result. The definition, statistics and Cohen's kappa score of each label is shown in Table 1.

### 3.2 Machine Learning Classifiers

We apply traditional Bag-of-Words [18] model to tokenize each reddit comment. On tokenizing to spare matrix, term frequency–inverse document frequency (TF-IDF) is used to adjust weight of each word in order to highlight the keywords of each comment. We will apply conventional machine learning algorithms to build a multi-class (3-class) classifier. The algorithms including naive Bayes, random forest, support vector machine with linear kernel [19], logistic regression and k-nearest neighbors. These conventional algorithms will be compared with deep learning models mentioned in this section.

Table 2: Performance of 5 machine learning classifiers and 2 deep neural networks on trangender sentiment analysis

| Algorithm | Accuray | Averaged AUC |
|---|---|---|
| Naive Bayes | 0.777 | 0.720 |
| Random Forest | 0.821 | 0.868 |
| Support Vector Machine | 0.832 | 0.842 |
| Logistic Regression | 0.829 | 0.851 |
| K-Nearest Neighbour | 0.764 | 0.767 |
| Convolutional Neural Network (CNN) | **0.861** | 0.834 |
| Long Short-Term Memory (LSTM) | 0.852 | **0.876** |

### 3.3 Deep Learning Models

We adopt convolutional neural networks (CNN) and recurrent neural networks to train a text classifier. Before feeding into the deep neural networks, each comment will be represented by a word2vec embedding [20] that is pre-trained on large scale social media data. The detailed configuration of CNN is listed as follows: the network is stacked with 3 convolutional layers with 128, 64, and 32 7-dimension filters; each convolutional layer is activated by a rectifier (ReLU) activation function, a global max pooling and a dropout layer (dropout rate is 0.2); a dense layer is added to the convolutional features to 3 classes. For RNN network, we stack 3 layer of long short-term memory (LSTM) [21] layers with 128, 64 and 32 neurons. Each layer is followed by a sigmoid activation function. Same embedding scheme as CNN is applied. For both deep neural networks, we employ Adam optimizer and binary cross-entropy loss function. Batch size is set to be 32. We train each model for 100 epochs with early stopping conditioned on the loss of development set.

### 3.4 Evaluation

We leave 20% of the entire dataset as the held-out test. Accuracy and averaged AUC across all 3 labels used to evaluate the performance of our models. For conventional classifiers, 5-fold cross validation is used to select best configurations. For deep neural networks, 10% of the training set is used as the development set to tune hyper parameters.

### 3.5 Implentation

All experiments are implemented using Python 3.6. Keras is used to design deep neural networks. Conventional machine leanring classifiers are built with scikit-learn package. We used a single GPU to expedite the training of deep neural networks.

## 4 Results

The annotation results are shown in Table 1. Among the 300 reddit comment, 72 are categorized as negative post, while 85 are annotated to be positive content. The rest 48% are considered as neutral sentiment. The Cohen's Kappa score is over 0.8 across all classes, which can be considered as a powerful annotation with high inter-rating agreement.

The performance of our NLP methods is shown in Table 2. Naive Bayes method serves as the baseline performance. All other models outperforms baseline by more than 5 percent on accuracy except K-nearest neighbour. Convolutional neural network yields the best accuracy of 0.861, followed by Long Short-Term Memory, which has an accuracy of 0.852. These 2 deep neural networks achieve better performance than all conventional machine learning classifier. LSTM also obtains an optimal AUC of 0.876, follwed by random forest, which has an AUC of 0.868. In short, LSTM yields the best performance when evaluated both on accuracy and AUC.

## 5 Discussion and conclusion

We visualize the single word's weights in logistic regression to discuss the important words that contribute to classification in Fig. 1. Top 20 words are visualized to a word cloud figure. We can see it is natural that *trans* and *transgender* are two words of the greatest importance. Besides this *dysphoria* is also heavily mentioned in the comments. The words: *question*, *wondering*, *anything* and *anyone* are also very important to the classification. These words may related to the depression and helpless of transgender community.

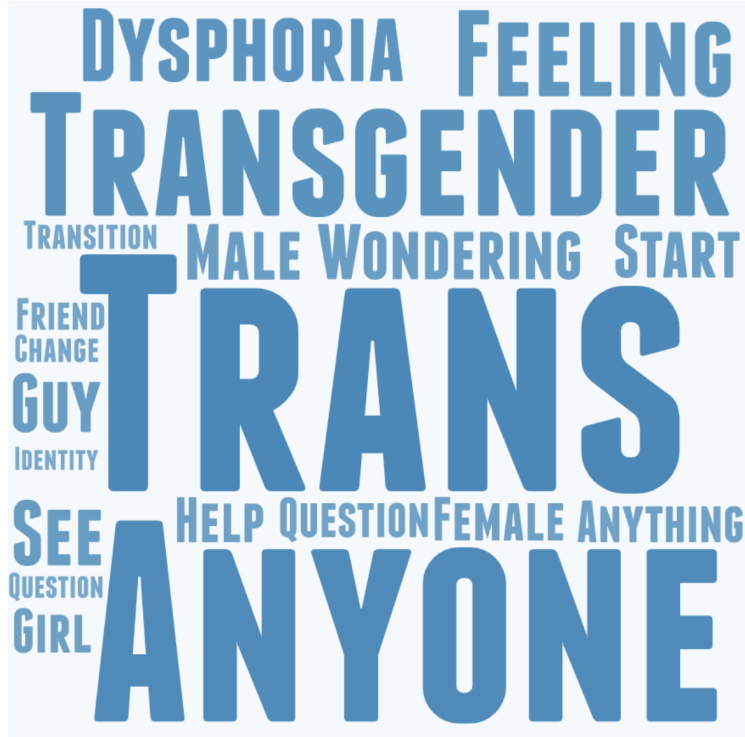

Figure 1: A word cloud of Top 20 important words contribute to classification

In terms of real case use, our classifier can identify negative sentiment from transgender people when they post on social media. This may help health providers understand the mental condition of transgender community better. It can be also used to detect extreme negative sentiments from these community and send alert to the friends, doctors or social media users themselves so that early intervention can be applied.

Our work also have following limitations. First, we only annotate 300 reddit comments which may not large enough for our NLP algorithms to build a solid automatic model. Second, we don't apply the most trending NLP algorithms such as graph convolutional networks [22] and transformer-based BERT model [23]. In the next step, we are planning to employ more powerful algorithms on a larger annotated dataset.

## 6 Author Contribution Statement

The research idea is discussed by all authors. Y.L. and Y.W annotate all 300 reddit comments. For those annotations without agreement between Y.L. and Y.W., Y.Z. is involved to give a final decision. Y.L., Y.W. and Y.Z. complete all experiments, error analysis and manuscript draft writing. Z.L. modifies the manuscript to a higher academic standard.